\documentclass[conference]{IEEEtran}
\ifCLASSOPTIONcompsoc
  \usepackage[nocompress]{cite}
\else
  \usepackage{cite}
\fi
\usepackage{fixltx2e}
\usepackage{url}

\usepackage{mathptmx} 
\usepackage{amssymb}  
\usepackage{amsmath}
\usepackage{cite}
\usepackage{graphicx}
\usepackage{amsmath}
\usepackage{algpseudocode}
\usepackage{algorithm}
\usepackage{algpascal}
\usepackage[T1]{fontenc}
\usepackage[utf8]{luainputenc}
\usepackage{amssymb}
\usepackage{lineno,hyperref}
\usepackage{epsfig} 

\begin{document}
%
\title{Line-Circle: A Geometric Filter for Single Camera Edge-Based Object Detection}

\author{\IEEEauthorblockN{Seyed Amir Tafrishi}
\IEEEauthorblockA{Department of Automation and \\
Control Systems, University of Sheffield,\\
Sheffield, UK\\
Email: amirtafrishi@yahoo.com}
\and
\IEEEauthorblockN{Vahid E. Kandjani}
\IEEEauthorblockA{Department of Information\\
Technology, University College of\\ Nabi Akram, Tabriz, IRAN\\
Email: esmaeilzadeh\_v@kish.sharif.edu}}


%


\maketitle

\begin{abstract}
This paper presents a state-of-the-art approach in object detection for being applied in future SLAM problems. Although, many SLAM methods are proposed to create suitable autonomy for mobile robots namely ground vehicles, they still face overconfidence and large computations during entrance to immense spaces with many landmarks. In particular, they suffer from impractical applications via sole reliance on the limited sensors like camera. Proposed method claims that unmanned ground vehicles without having huge amount of database for object definition and highly advance prediction parameters can deal with incoming objects during straight motion of camera in real-time. Line-Circle (LC) filter tries to apply detection, tracking and learning to each defined experts to obtain more information for judging scene without over-calculation. In this filter, circle expert let us summarize edges in groups. The  Interactive feedback learning between each expert creates minimal error that fights against overwhelming landmark signs in crowded scenes without mapping. Our experts basically are dependent on trust factors' covariance with geometric definitions to ignore, emerge and compare detected landmarks. 
The experiment for validating the model is taken place utilizing a camera beside an IMU sensor for location estimation.
\end{abstract}

\section*{Nomenclature}
\addcontentsline{toc}{section}{Nomenclature}
\begin{IEEEdescription}
	\item[$\lambda$] Collector for grouping landmarks.
	\item[$\psi$] The ignorance parameter.
	\item[$E_n,\;E_r$] Normal and rebel edges matrices.
	\item[$C_n,\;C_r$] Normal and rebel collected landmark matrices in Circle expert.
	\item[$\alpha$] Scaling parameter to determine rebel edges.
	\item[$L$] Edge location on the captured frame.
	\item[$BS,BL$] Boundary size of $\lambda$ and $E_n$.
	\item[$DL$] Deviation level of rebel edges.
	\item[$R$] Radius of given scaling/temporary parameters.
	\item[$Ty$] Type of found ignorance region
	\item[$Tr$] Trust factor.
	\item[$\beta_q$] Relevant angle of edge or experts.
	\item[$V_q$] Velocity magnitude of edge or experts.
	\item[$O$] Existing origin point for rebel landmarks.
	\item[$\delta$] Governing error matrix.
	\item[$Edge$] Current frame's edges.
\end{IEEEdescription}

%
\IEEEpeerreviewmaketitle

\section{Introduction}
In the most of autonomous driving systems, it is important to detect objects along the direction of moving vehicle. Mobile robots have limited source of power and require fast responses during their motion. Additionally, it is an urge to reduce resources and sensors during active times \cite{Visiononl2014}.
There have been some studies on object detection and mapping to enhance the performance and decrease the computation  \cite{Visiononl2014}-\cite{TLD2012}. These detectors are using real-time data to process during their activation states. 

Different detection methods have been studied to create accurate and fast analysis over the surrounding. Eade and Drummond proposed a machine learning corner detection method to be used as the first monocular SLAM systems capable of operating in real-time \cite{Rosten2006}. As a classic approach for understanding the environment the stereo-based model was firstly designed by Murray and Little for SLAM operations \cite{Murray2000}. Other detection methods were studied related to the point analysis \cite{Sand2008} and image contouring \cite{bibbyeccv08}. Recently, semi-dense monocular SLAM with integration of color was applied to determine the surrounding objects at MIT \cite{PillaiRSS2015}. Lastly, an overall review was presented by Sun et al about localizing other vehicles on the road \cite{VehicleREview2006}. 

Monocular SLAM with corner detection as the most practical and advance method, was analyzed to have better solutions for overconfidence and dealing with high computations related to detected landmark. Lui and Drummond proposed a new system for constant time monocular SLAM that uses only 2D measurement and takes the image graph with sparse pairwise geometric. There were some improvements such as no global consistency and bundle adjustments but the system was based on multi-camera perspective and it was also dependent on a reference image \cite{Tomy2015ICRA}. As a alternative, Kalman filter reduction was operated for SLAM problems for using bundle adjustment by sparse matrix and double windowing method \cite{TommyIROS2015}. Gamage and Drummond tried to decrease dimension of the co-variance matrices of camera and landmark position. Although it minimized some nonlinearities that creates inconsistency in EKF, it was not able to deal with overwhelming landmarks.
A requirement for solely image-based filter as a superior method rather than a classic Kalman filter has been remained unsatisfactory. Vision dependent applications require filter that is able to do real-time computation reduction within diverse machine learning evaluation layers.

In this paper we constructed a filter for solely 2D image utilization about object detection. In following work, via using a real-time captured data parallel to the internal measurement unit (IMU), a scene is utilized to predict possible objects that are passing or moving towards our vehicle. Additionally, by using velocity and vehicle dynamics, this filter optimizes the data collection from existing edge detection methods or event-base camera \cite{EVENTBASECAMERA} to ignore/concentrate on particular landmarks. Beside these capabilities, with certain inspiration from estimators namely Kalman Filter, errors are minimized to prevent any wrong or integrated uncertainties of detected landmarks.

\section{Line-Circle (LC) Filter}
\begin{figure}[t!]
\centering
\includegraphics[width=3.2 in]{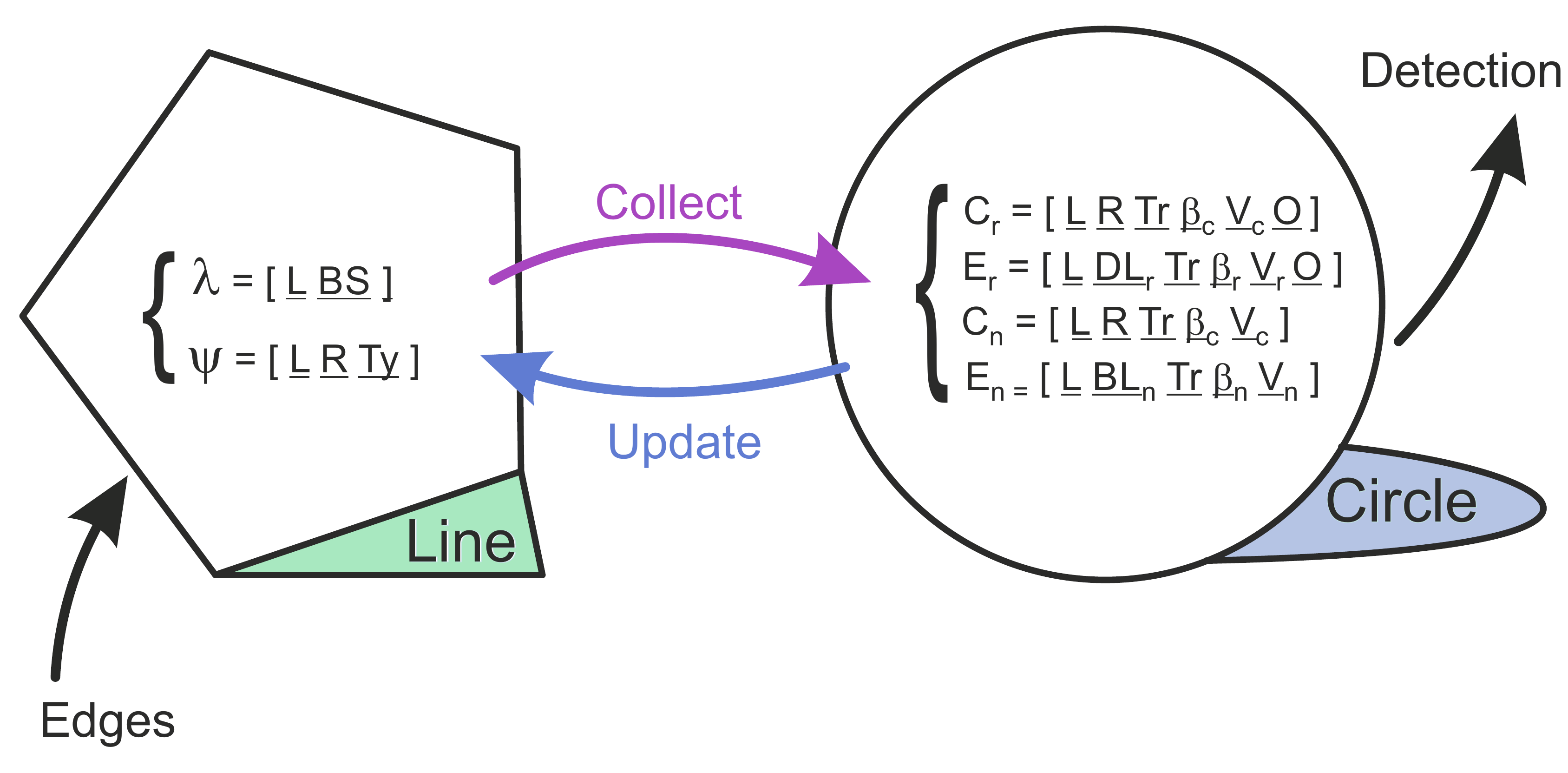}
\caption{The map of geometric filter.}
\label{Fig:AlgorithmFunctioningMAP}
\end{figure}

The filter is proposed in a way that the mainframe system have its evaluation from surroundings in real-time. The algorithm basically understands and creates the object locations with its past information and currently obtained data from camera and IMU sensors. The novelty of approach stands about its multi-level analysis on captured edge from corner detection method \cite{Rosten2006,4674368}. Fundamentally, this method carries its environmental analysis with filtering the data in each state of experts (line and circle) to update the previous one for future incoming frames beside next expert. The overall frame studies have been mapped like an interactive transitional-base principles with main parameter flow for detection, learning and tracking the incoming detected corners [see Fig. \ref{Fig:AlgorithmFunctioningMAP}].

Scaling and temporary parameters are two groups in this approach. In former, $\lambda$ is responsible for grouping the edges with $L$ location and $BS$ boundary size. $\psi$ is a feedback parameter for $\lambda$ in which it takes the information from circle to create ignorance regions via $L$, $R$ and $Ty$ as the location, radius of ignorant and a flag for ignoring geometry type. $\psi$ avoids trusted overloading edges involvement in flow of analysis so it helps the robot to process faster and concentrate more on the essential locations. Parameters in latter are temporary information carriers from previous frames. $E_n$ and $E_r$ are the parameters responsible for estimated edges to determine the their properties. Beside these parameters in Circle Expert, $C_n$ and $C_r$ will assist the system to collect edges with relevant properties to prepare it for understanding objects without requirement to know their formations. Each parameter has a collection series of information about $L$ location, $R/BL_q$ size of space, $Tr$ trust factor beside $\beta_q$, $V_q$ as the angle and velocity of corresponding motion parameters. $O$ stands for the originated landmark for rebel parameters.

\begin{figure}[t!]
\centering
a)\includegraphics[width=2.2 in]{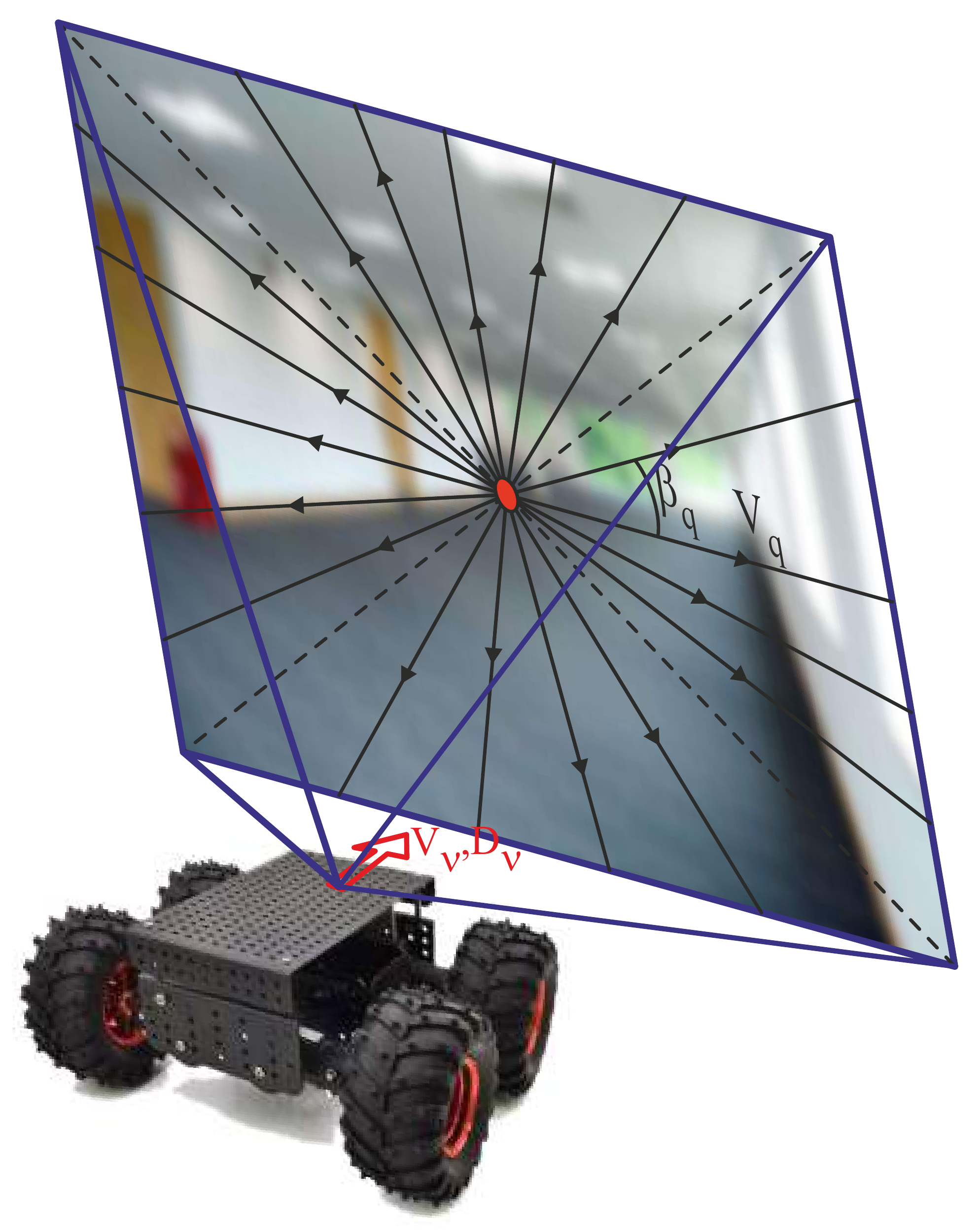}
b)\includegraphics[width=2.2 in]{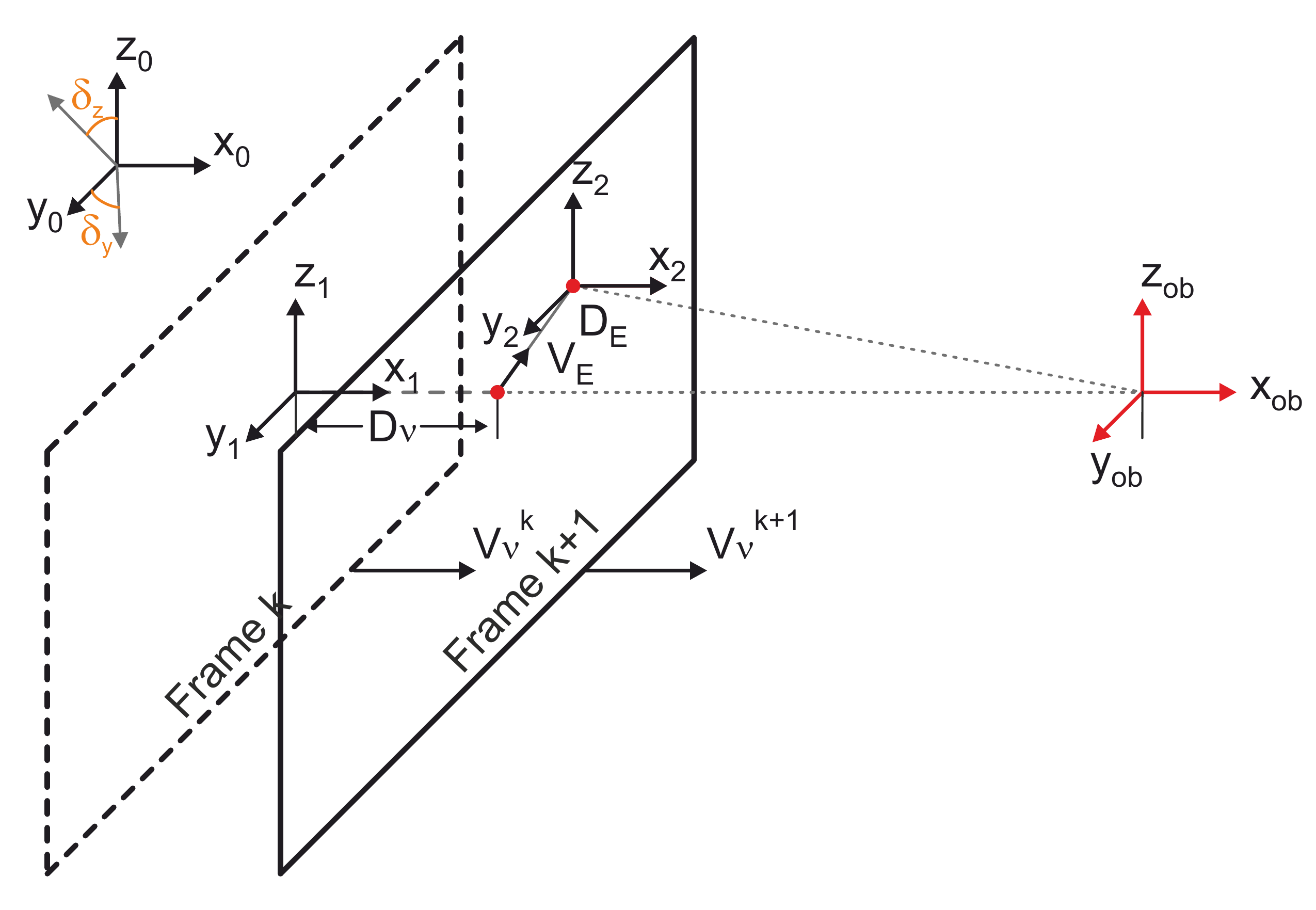}
\caption{Kinematic analysis of locomotion with orientated vector field in frame. a) The general view b) Kinematic analysis of edges motion respect to frame and corresponding object.}
\label{Fig:VectorField}
\end{figure}

We have four transitions in this machine learning algorithm. Each transition carries detection, learning and tracking applications working with feeding data to experts. Therefore, It explains our dynamic modeled approach that works in integration without sole reliance on a pyramidal supervisory \cite{Bouguet00pyramidalimplementation}. In first step (base transition), the camera collects the detected edges and groups/omits them with relative feedbacks from previous steps in Line Expert. Next, collection transition happens to categorize the edge groups with reliance on their behaviors to create circles with taken feedback. As a final stage, the estimated variables are presented to have classified data from truly existing objects in Circle expert. The basic kinematic analysis beside parameter definition are as Fig. \ref{Fig:VectorField} where mobile robot follows a straight line.

As an important fact, the flow distribution and intensity of vector field is related to plain angle, robots rotation and locomotion velocity. In this practice [see Fig. \ref{Fig:VectorField} case b], the two frame information is used where the collected information about vehicles $D_\nu$ and $V_\nu$ are compared mathematically for deriving the relevant $D_e$ and $V_e$ of relevant edge. Because in this work, it is assumed that camera just moves in straight line, any angular motion in z and y axes are feed as error $\delta_{z,y}$.

As a simple explanation, Line obtains the edges and classifies them with locational boundaries. Next, Circle concentrates on decision making of the most suitable edge locations as well as re-grouping the edges with velocity and other characteristics. These experts interact with machine learning procedure to update their classifications interactively (e.g. trust evaluators and created errors).
\subsection{Line Expert}
Line expert contains straight forward computation. Algorithm \ref{Algo:line} shows the general order of its work. 
\alglanguage{pseudocode}
\begin{algorithm}[t!]
\caption{Line Expert}
\begin{algorithmic}[1]
\Procedure{Line}{$\lambda, \psi$, $Edge$}
\While {All edges are checked in $Edge$}
\If {The edge is  within circular space of $\psi$}
\State Omit the variable from $Edge$
\EndIf
\EndWhile
\State Group the $Edge $ by $\lambda$\\
\Return $Edge$
\EndProcedure
\end{algorithmic}
\label{Algo:line}
\end{algorithm}
The first expert with using the instructed ignored regions $\psi$ and the collected $\lambda$ for proper boundary choices in different regions of frame, groups the edges in $Edge$ matrix. Also, newly appeared unmatched landmarks are added as well. $\lambda$ detects groups with a given equation:
\begin{equation}
\sqrt{(L_{\lambda_{x}}-Edge_x)^2+(L_{\lambda_{y}}-Edge_y)^2}\; < \;BS_{\lambda}
\label{BSsizecal}
\end{equation}
$\psi$ removes the edges via including the expressed regional constraints for different $Ty$: 
\begin{equation}
\begin{cases}
\begin{array}{c}
0\;\;\;\;\;\;\;\;\;\;\;\;\;\;\;\;\;\;\;\;\;\;\;\;\;\;\;\;\;\;\;\;\;\;\;\;\;\;\;\;\;\;\;\;\;\;\;\;\;\;\;\;\;\;\;\;\;\;\;\;\;\;\;\;\;\;\;\;\;\;\;\;\;\;\; Ty\;=\;0\\
\pi R_{\psi}^2\;\;\;\;\;(L_{\psi_x},L_{\psi_y})\;\;\;\;\;\;\;\;\;\;\;\;\;\;\;\;\;\;\;\;\;\;\;\;\;\;\;\;\;\;\;\;\;\;\;\;\;\;\;\;\;\;\;\;\;Ty\;=\;1\\
R_{\psi_x}R_{\psi_y}\;\;\;\;\;(L_{\psi_x},L_{\psi_y})\;\;\;\;\;\;\;\;\;\;\;\;\;\;\;\;\;\;\;\;\;\;\;\;\;\;\;\;\;\;\;\;\;\;\;\;\;\;\;\;Ty\;=\;2
\end{array}
\end{cases}
\label{Psijun}
\end{equation}
By having $Ty\;=\;1$, equation (\ref{Psijun}) covers circular area or quadrangles area with $Ty\;=\;2$. Lastly, algorithm returns to the ordered $Edge$ with changed form. It is important to note that the detected corners are in coupled locations in every two columns of matrix $Edge$. 
\subsection{Circle Expert}
This essential expert "Circle" tries to learn and determine the detected landmark to do tracking better. This stage transforms static edge groups $\lambda$ to dynamical patterns. This expert helps the information be more practical specially for future usages at mapping tasks.  
\begin{figure}[t!]
\centering
\includegraphics[width=3.2 in]{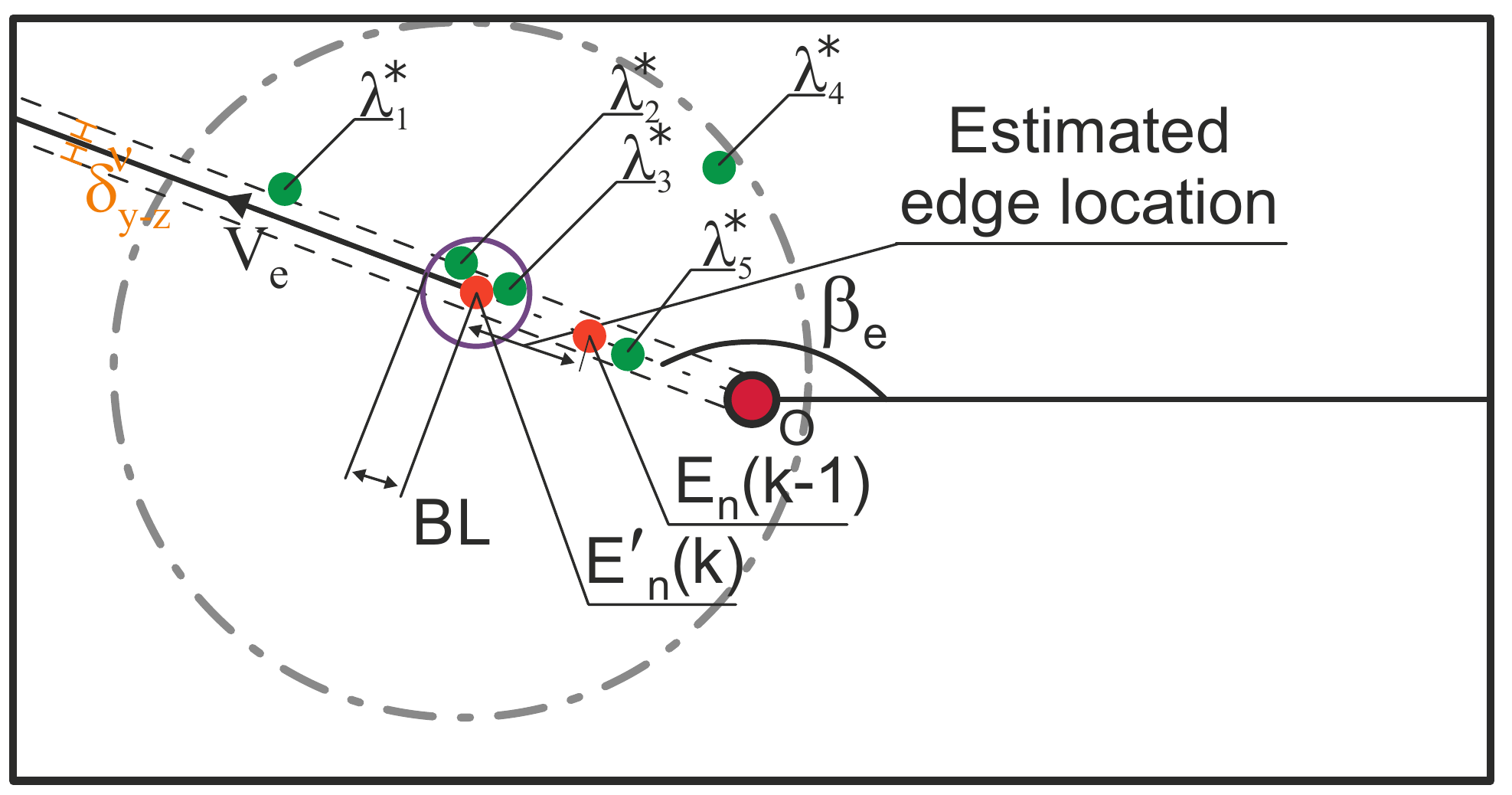}
\caption{The obtained edge's classifications with current information of $E_n$.}
\label{Fig:EdgeEvaluationforlocationesti}
\end{figure}
The principle definitions for edges locations are assumed as illustrated Fig. \ref{Fig:EdgeEvaluationforlocationesti}. By considering the red dot in the center of locomotion with aligned plain, detected landmarks can be located with five basic scenarios $\overset{5}{\sum} \lambda^*$. $V_e$ in here is the velocity of edge. It is important to note that the properties of $V_e$ and $\beta_e$ (edge angle) are for the edges that mostly are passed objects. This is why the velocity is aligned with center of frame O. For case of rebel edges, the center O will vary depending on the coming origin. The flow of normal edges vectors are linear since we assumed robot is following a direct locomotion without rotations. These models will be later explained in Normal Edge Circling section. The dashed double lines express the error span $\delta^{\nu}_{y-z}$ depending on rotational error. However, $BL$ is learning and changing related to factors through the locomotion. 
\subsubsection{Trust Evaluators}
Before the detailed explanation of edge classifications, the trust ($\forall  Tr \;\in \; N$) as the main role in machine learning part of the filter is ranked in three values:

\textbf{$Tr_s$:}
It is the standard trust for representing the values, if trust level of parameter is equal or greater, the parameter is highly true with minor errors. These parameters with this specification are able to determine the edge/circle locations with high accuracy. 

\textbf{$Tr_{cr}$:} It is a critical magnitude in general trust definition. When trust value is lower than this, parameter is in the risk of deleting because of reasons like passed object's edges or wrong estimations.

\textbf{$Tr_{max}$:} The maximum value in trust that constrains the parameters for preventing overconfidence and using evaluating parameters such as $\psi$.

Each of these ranked trusts, lets the filter properly eliminate, re-coordinate or combine the edges or circle groups. Moreover, every newly created edge/circle has 
\begin{equation*}
Tr=\frac{1}{2}[Tr_{cr}+Tr_{s}]
\end{equation*}
trust value. 
\subsubsection{Edge Classifications}
As it was explained, before giving the estimation definitions [see Fig. \ref{Fig:EdgeEvaluationforlocationesti}], the trust function is a primary variable that in each frame depending on the obtained evaluations through classifications, we consider how far the existing reliability can be implemented for our future utilizations on the captured scenes. These classification let us determine new candidate, low accurate detected, rebel and normal flowing edges from each other.

Therefore, $\lambda^*$s as all possible estimated landmark locations are expressed with below classifications:

$\bullet$ \textbf{$\lambda^*_1$}: The edge exceeds $BL$ region for estimated one but exists in error span. This parameter will be added as an 	extra normal edge inside the included circle but the trust value of estimated edge will be upgraded as -1.

$\bullet$ \textbf{$\lambda^*_2$}: By being in $BL$ area and satisfying the error span this will be normal edge which is the suitable candidate for evaluation. Also, the trust of corresponding edge will be upgraded as +1.

$\bullet$ \textbf{$\lambda^*_3$}: This edge despite the satisfaction in $BL$ region was failed from $V_e$ and $B_e$ characteristics. Furthermore, It will be required to upgraded with error inclusion. Nevertheless, the trust value for estimated edge will be upgraded as -1.

$\bullet$ \textbf{$\lambda^*_4$}: With breaking all the laws depending on the $V_e$, $\beta_e$, $BL$ and error span, it will be considered as rebel edge or new normal landmark depending on the incoming frame rebel edge evaluations. The trust value of estimated edge will remain the same. A new normal edge with average standard and critical trusts will be constructed.

$\bullet$ \textbf{$\lambda^*_5$}: Despite exclusion from estimated edge $BL$, it is in the flow line of error span within $BL$ of previous edge itself. It is assumed as rebellious landmark which leads rebel edge category for further evaluations. The trust of normal edge will be decreased and it will be estimated as $E'_n(k)$ but data will be carried to $\alpha$ for validation of possible $E_r$ existence.
\begin{figure*}
	\centering
	\includegraphics[width=2 in]{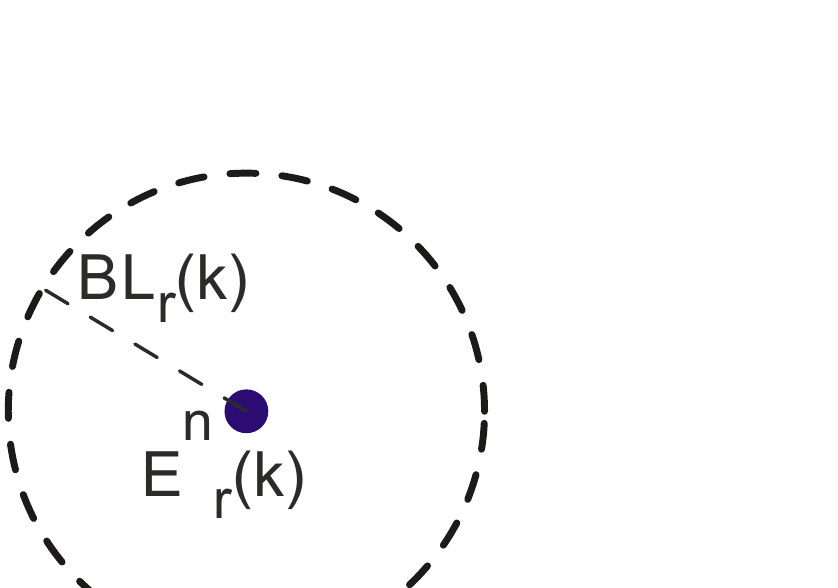}
	\includegraphics[width=2 in]{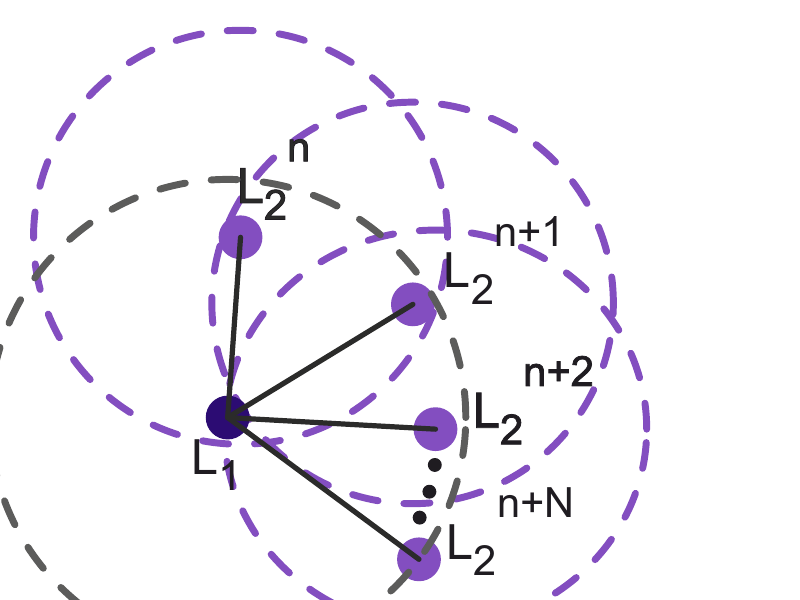}
	\includegraphics[width=2 in]{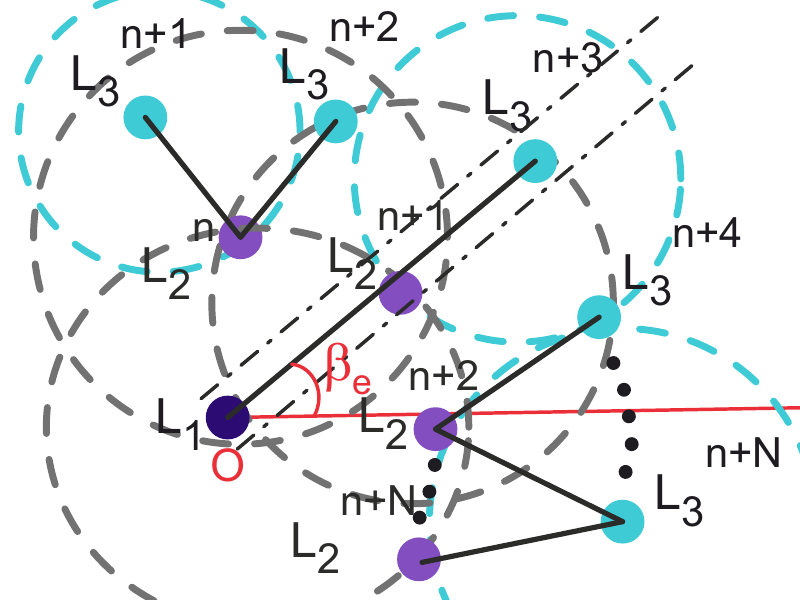}
	\newline
	\space\space  Frame k\space\space\space\space\space\space\space\space\space\space\space\space\space\space\space\space\space\space\space\space\space\space\space\space\space\space\space\space\space\space\space\space\space\space\space\space\space Frame k+1 \space\space\space\space\space\space\space\space\space\space\space\space\space\space\space\space\space\space\space\space\space\space\space\space\space\space\space\space Frame k+2 \space\space\space\space\space\space\space\space\space
	\caption{Rebel landmarks orientation detection through the three key step frames. The dashed gray lines are the previous steps boundary layers, the dot-dashed lines are the error span variation's space. The remaining colorized dashed circles are corresponding standard BL size for obtained marks.}
	\label{Fig:EDgerebellers}
\end{figure*}

In initialization of $E_n$, landmarks are placed with no estimation which edge carries its velocity as the vehicle velocity and angle is defined depending on the frame center O beside initial $BS_0$. Later on, the normal edges $E_n$ are detected from (\ref{BSsizecal}) and error span as:
\begin{equation}
\frac{|-(Edge_x-IC_x)+m_e(Edge_y-IC_y)|}{(1+m_e^2)^{\frac{1}{2}}} < \delta^{\nu}_{y-z}
\end{equation}
Where $IC_{x,y}$ and $m_e$ are the image center locations and slope of $N_{th}$ $E_n$ respect to frame center. $E_n$ is estimated with beneath formulas for edges location, velocity and boundary layer:
\begin{multline}
\begin{split}
&L_{E_n}(k)=\frac{(Tr_{E_n}(k-1)-Tr_{cr})L'_{E_n}(k-1)+L_{Edge}(k)}{(Tr_{E_n}(k-1)-T_{cr})+1}\\
&V_{E_n}(k)=|V_{\nu}(k)\\
&\pm\frac{(L_{{E'_n}_{x}}(k-1)-Edge_x)^2+(L_{{E'_n}_{y}}(k-1)-Edge_y)^2}{t_f}|\\
&BL_{E_n}(k)=\frac{1}{2}[\frac{|V_{\nu}(k)-V'_{E_n}(k-1)|}{Corr(Edge^n_{group })}+BL_{E_n}(k-1)]\\
\end{split}
\label{Eq:EnEstimation}
\end{multline}

Because it is assumed model is able to recover the error $\delta^{\nu}_{y-z}$, $\beta_{E_n}$ is remained unchanged in straight camera motion. Also, as $\beta_{E_n}$ has to be estimated in the first step to determine corresponding edge,  the error velocity is using $V'_{E_n}(k-1)=\frac{1}{2}[V_{E_n}(k-1)+Vv]$. $t_f$ is the spending time from previous frame till the current Circle expert run.

\alglanguage{pseudocode}
\begin{algorithm}[t!]
	\caption{Circle Expert, Edge Estimation Part}
	\begin{algorithmic}[1]
		\Procedure{Circle}{$Edge,C_n,C_r,E_n,E_r,\psi,\delta,V_\nu,D_\nu$}
		\If {$E_n$ = $\varnothing$} \Comment{Edge matching}
		\State Initialize $E_n$ with $Edge$
		\Else
		\While {All the edges in $E_n$ are checked}
		\While {All the $Edge$ groups are checked}
		\State Find matching landmark group of $Edges$
		\If {$N^{th}$ $E'_n$ satisfies $\lambda^*_{2,3}$}
		\State Find nearest $Edge$ in boundary
		\State Update  $N^{th}$ $E_n$
		\State Remove landmark from $Edge$
		\ElsIf {$N^{th}$ $E'_n$ satisfies $\lambda^*_{5}$}
		\State Update $N^{th}$ $E_n$
		\State Apply rebel detection with $\alpha$
		\State Remove landmark from $Edge$
		\ElsIf {No $E'_n$ match}
		\State $N^{th}$ $E_n$ $\leftarrow$ $N^{th}$ $E'_n$
		\State Update $N^{th}$ $E_n$ remaining param.
		\EndIf
		\EndWhile
		\EndWhile
		\If {$E_r$ = $\varnothing$}
		\Else
		\While {All the edges $E_r$ are checked}
		\While {All the $Edge$ are checked}
		\State Find landmark from $Edges$
		\If {$N^{th}$ $E'_r$ satisfies $\lambda^*_{r}$}
		\State Update $N^{th}$ $E_r$
		\State Remove landmark from $Edge$
		\ElsIf {No $E'_r$ match}
		\State $N^{th}$ $E_r$ $\leftarrow$ $N^{th}$ $E'_r$
		\State Update $N^{th}$ $E_n$ remaining param.
		\EndIf
		\EndWhile
		\EndWhile
		\EndIf
		\EndIf
		\State Add left $Edge$ as new $E_n$ ($\lambda^*_{1,4}$) \Comment{Edge matching}
		
		\Return $E_n$, $E_r$, $\alpha$
		\EndProcedure
	\end{algorithmic}
	\label{Algo:Circle1}
\end{algorithm}

For determining the rebel edges $E_r$ detections, we proposed line tracking model that tries to eliminate these edges with N frame steps [see Fig. \ref{Fig:EDgerebellers}]. By presenting the minimum $N=3$ frame per sec, we sum the detected landmarks by following their left marks from previous steps. We apply the model with considering underneath specifications:

$\bullet$ Due to unexpected motion in the rebellious landmarks for consecutive frames, tracking will be a  non-linear motion with a certain deviation in each frame.

$\bullet$ The number of frame to evaluate the reliability of existing detection to rebel landmarks will be 3.

$\bullet$ The relation of the connected landmarks are deleted after success/fail in three frame analysis of $\alpha$. 

The $\alpha$ scaling parameter basically works as following: 
\begin{equation}
\alpha \varpropto \{ \overset{3}{\sum}L_n, \ell \} 
\end{equation}
Where $L_n$ is the located edges positions on frame and $\ell$ is the number of frames that the involved edges are checked. However, this study includes the error span as well in this formulation. The initial values for $E_r$ is constructed as following:
\begin{multline}                                   
\begin{split}                                      
&{DL_{E_r}}_0= \hat{L}_{3-1} - \hat{L}_{2-1}\\             
&{V_{E_r}}_0=\frac{L_3-L_2}{t_f}\;\;\;\;\;\;\;\;\;\;\;\;\;\;\;\;\;\;\;\;\;\;\;\;\;\;\;\;\;\;\;\;\;\;\;\;\;\;\;\;\;\;\;\;\;\;\;\;\;\;\;\;\;\;\;\;\;\;\;\;\;\;\;\;\;\;\;\\
&{\beta_{E_r}}_0=\hat{L}_{3-1}\\
&{O_{E_r}}_0=L_1
\label{Eq:Intialrebelpara}
\end{split}
\end{multline}
Where, for a example, $\hat{L}_{3-1}$ presents angle of a constructed line from point 3 with respect to 1. For existing rebel landmarks the parameters $L$, $V_r$ are estimated 
as equations (\ref{Eq:EnEstimation}),(\ref{Eq:Intialrebelpara}) except $DL$ is updated as:
\begin{equation}
{DL_{E_r}}(k)={DL_{E_r}}(k-1)-[{\beta_{E_r}}(k)-{\beta_{E_r}}(k-1)-{DL_{E_r}}(k-1)]
\end{equation}
Addition to these, the ${\beta_{E_r}}(k)$ is just calculated from last detected edge respect to rebel edges origin $O$. These possible outcomes help us for having the detection and learning to be smooth and we would have similar appliance of negative and positive classified examples \cite{TLD2012}. In contrast, we use gradual machine-learning with disappearing pattern via truth definition. These assumptions with designed constraints and dynamic parameters can be seen as the main functioning point in the Circle expert. Algorithm \ref{Algo:Circle1} is showing the main execution codes in preparing the rebel and normal circles.
This designed algorithm is divided to three main parts. First part detects and compares the $Edge$ with $E_n$ group. Next, after elimination, remaining unmatched landmarks are carried to second part for $E_r$ rebel edges study. The latest $E_n$ and $E_r$ matrices are  compared under giving constraints to develop the latest circles.
\subsubsection{Normal Edge Circling}
\begin{figure}[t!]
\centering
\includegraphics[width=2.4 in]{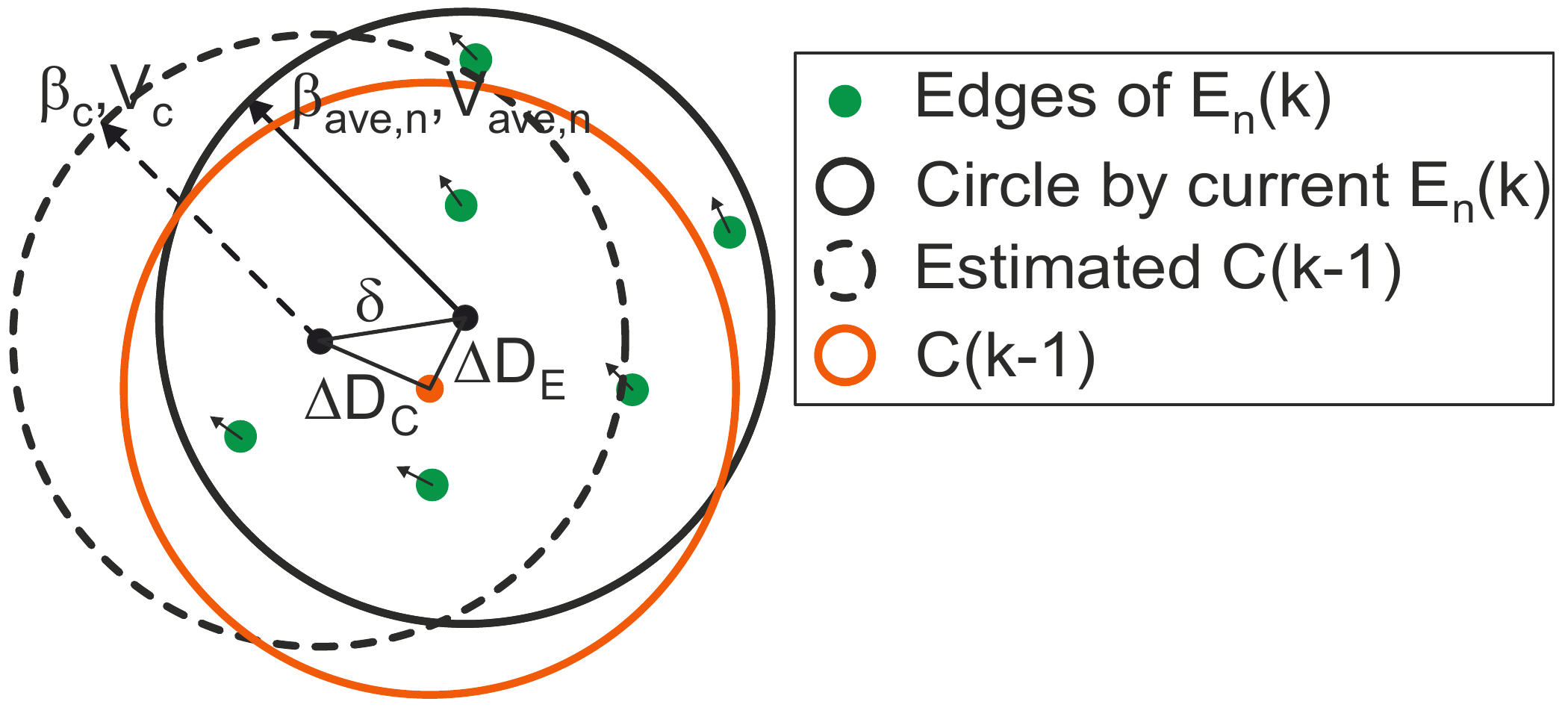}
\caption{Normal edges circle modeling.}
\label{Fig:Normaledge1}
\end{figure}

These landmarks are following the flow of vector field and group depending on their direction and velocities. To match these models, 
Fig. \ref{Fig:Normaledge1} is considered to form the circles.

\begin{algorithm}[t!]
	\caption{Circle Expert, Circle Estimation Part}
	\begin{algorithmic}[1]
		\Procedure{Circle}{$Edge,C_n,C_r,E_n,E_r,\psi,\delta,V_\nu,D_\nu$}
		\If {$C_n$ = $\varnothing$ \& $E_n$ $\neq$ $\varnothing$ } \Comment{$C_n$ matching}
		\State Initialize $C_n$
		\Else
		\While {All the circles in $C_n$ are checked}
		\While {All the edges in $E_n$ are checked}
		\State Group $E_n$ 
		\If {Grouped $E_n$ satisfies normal edge aligned $N^{th}$ $C'_n$}
		\State Update $N^{th}$ $C_n$
		\ElsIf {Grouped $E_n$ satisfies normal edge deviated $N^{th}$ $C'_n$}
		\State Update $N^{th}$ $C_n$
		\EndIf
		\EndWhile 
		\EndWhile
		\State Update not matched $C_n$ with $C'_n$
		\State Create new $C_n$ from left $E_n$
		\EndIf \Comment{$C_n$ matching}

		\If {$C_r$ = $\varnothing$ \& $E_r$ $\neq$ $\varnothing$ } \Comment{$C_r$ matching}
		\State Initialize $C_r$
		\Else
		\While {All the rebel circles in $C_r$ are checked}
		\While {All the edges in $E_r$ are checked}
		\State Group $E_r$ 
		\If {Grouped $E_r$ satisfies rebel circle aligned $N^{th}$ $C'_r$}
		\State Update $N^{th}$ $C_r$
		\ElsIf {Grouped $E_n$ satisfies rebel circle deviated $N^{th}$ $C'_r$}
		\State Update $N^{th}$ $C_r$
		\EndIf
		\EndWhile 
		\EndWhile
		\State Update not matched $C_r$ with $C'_r$
		\State Create new $C_r$ from left $E_r$
		\EndIf \Comment{$C_r$ matching}
		
		\If {Any $C_n$'s $Tr$ satisfies $Tr_{max}$}
		\State Update $\psi$ 
		\State Refresh $Tr$
		\EndIf
		\State Update $\lambda$ by $C_n$ and $C_r$ \\
		\Return $C_n$,$ C_r$, $\lambda$, $\psi$
		\EndProcedure
	\end{algorithmic}
	\label{Algo:Circle2}
\end{algorithm} 
Basically, kinematic estimation over the edges helps prepare the semi-autonomous system to evaluate its form with dependency on grouped edges in $Edge$. In this analysis, estimated circle from previous data and overwritten circle with obtained estimated edges are involved. This property is applied while it is within $BS_\lambda$ region. For grouping $E_n$s depending on their angle and velocity for both initialization and circle comparison below formulas are applied. 
\begin{multline}
\Bigg\{
\begin{split}
&(\beta_{{E_n}_{ref.}} - \varepsilon_{\beta} )< \beta^i_{E_n} < (\beta_{{E_n}_{ref.}} + \varepsilon_{\beta}) \;\;\;\;\;\;\;\;\;\;\;\;\;\;1<i\leq M\\
& |V^i_{E_n}| \leq \varepsilon_{V} V_{\nu} \\
\end{split}
\label{EdgeEncompare}
\end{multline}
The $\varepsilon$ depending on the required accuracy and distributional numbers can be carried to find proper constraints. Next, to compare the existing $C_n$, we have:
\begin{multline}
\begin{split}
&(\frac{\overset{M}{\sum} \beta_{E_n}}{M+1}- \varepsilon_{\beta}) < \beta^i_{{C_n}}  <( \frac{\overset{M}{\sum} \beta_{E_n}}{M+1} + \varepsilon_{\beta}) \;\;\;\;\;\;\;\;\;\;\;\;\;\;\;\;\;\;\;\;\;\;\;\;\;\;\;\;\;\;\;\;\;\;\\
& \frac{|\overset{M}{\sum} V_{E_n}|}{M+1} \leq \varepsilon_{V} V^i_{C_n}\\
\end{split}
\label{Eq:EdgeCirclenormalcomparee}
\end{multline}
Before applying these comparators, a weighting function takes place to evaluate the percentage of involvement about right circle in the these collected $E_n$.
\begin{equation}
\% \{(L^i_{E_n}-L_{C_n}) < R_{C_n}\} < \% Cte
\label{Eq:percentinvolve}
\end{equation}
Back to the circle estimation, the $\triangle D_C$ and $\triangle D_E$ are calculated with simple trigonometric formulations, and proportional to the $\delta$ and trust value of circles, the reliable circle is optimized. As an important case, while circle's angle is aligned with average grouped $E_n$ with defined $\varepsilon_{\beta}$, the corresponding circle is upgraded with +1 trust. However, if angle is deviated despite major inclusion of edges in the circle, it is updated with decreasing trust. 


\subsubsection{Rebel Edge Circling}
\begin{figure}[t!]
\centering
\includegraphics[width=2.4 in]{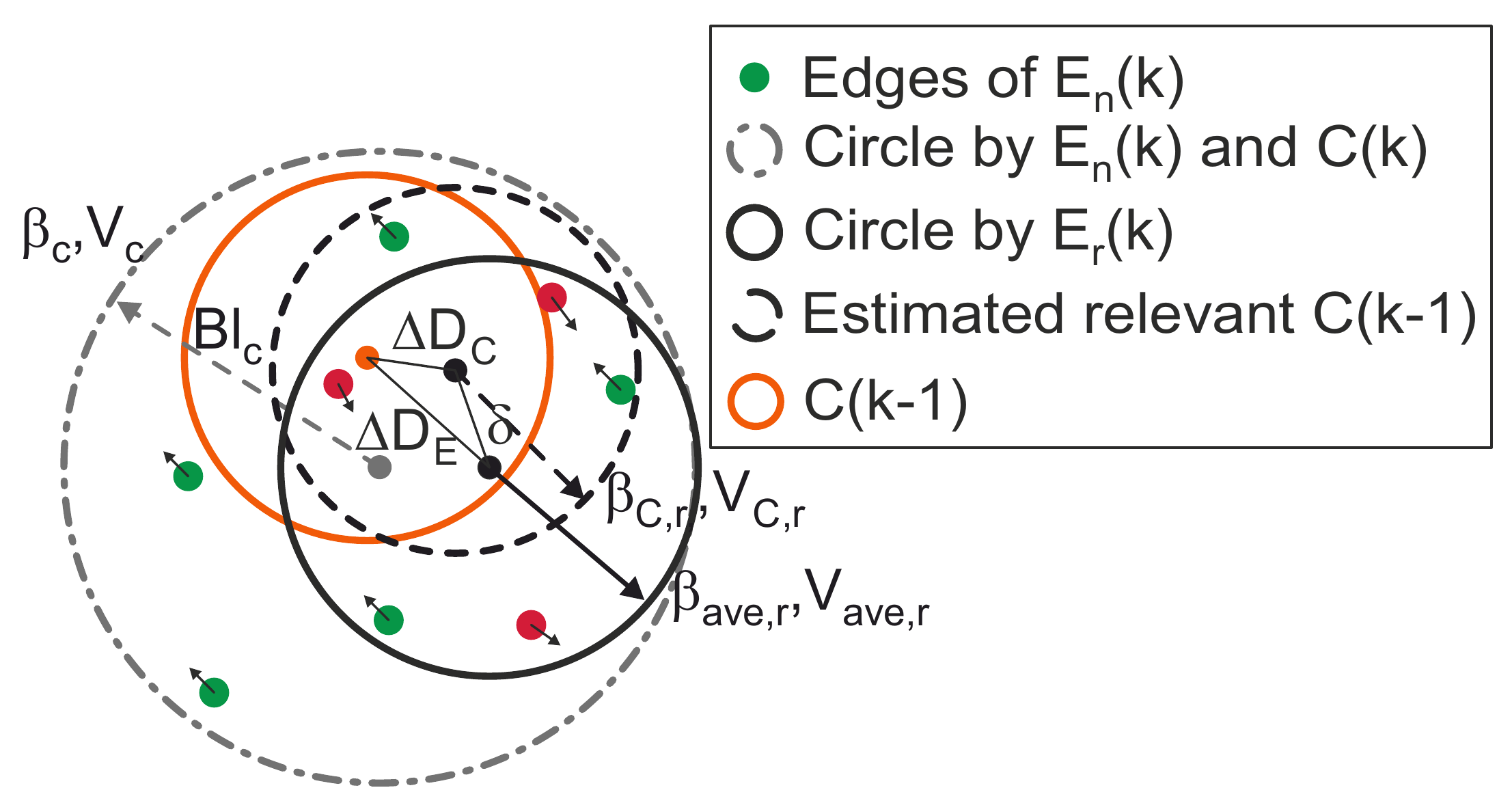}
\caption{Rebel edges circle modeling, Note: Red dots are detected rebel edges.}
\label{Fig:Normaledge2}
\end{figure}

Conspicuously, the most of objects won't have constant flow on the frame. These objects normally will be the ones that have potential to come toward or critically approach the captured camera. Therefore, rebellious detected edges will have most importance when it comes to SLAM or object avoidance applications. The rebel landmarks in the worst case scenario can be appeared within the normal edge circles [see Fig. \ref{Fig:Normaledge2}]. To evaluate these formats, the previous available similar edges are considered with specific dedication to the $\beta_{C_r}$ and $V_{C_r}$. Furthermore, it must be considered these edges are the hardest ones since they are not following vectors field flow when they are having a displacement on frame. Thus, it is not able to carry out with accurate reliance on the next frame similar velocity edges in surrounding area of located ones. Therefore, the equation (\ref{EdgeEncompare}) is improved as:
\begin{multline}
\Bigg\{
\begin{split}
&(\beta_{{E_r}_{ref.}} - \varepsilon_{\beta}) < \beta^i_{E_r} < (\beta_{{E_r}_{ref.}} + \varepsilon_{\beta}) \;\;\;\;\;\;\;1<i\leq M\\
& |V_{{E_r}_{ref.}}| \leq (V^i_{E_r}+\varepsilon_{V} V_{\nu}) \\
\end{split}
\label{Eq:EdgeErcompareER}
\end{multline}
Velocity constraint is largely dependent on rebel edges estimated velocities. Before estimation, we have to track the circle in which is using the same technique (\ref{Eq:percentinvolve}). Next, the comparator is evaluating the grouped rebel edges with each $C_r$ as following:
\begin{multline}
\begin{split}
&(\frac{\overset{M}{\sum} [\beta_{E_r}+DL_{E_r}]}{M+1} - \varepsilon_{\beta} )< \beta^i_{{C_r}}  < (\frac{\overset{M}{\sum} [\beta_{E_r}+DL_{E_r}]}{M+1}  + \varepsilon_{\beta})\;\;\;\;\\
& \frac{|\overset{M}{\sum} V_{E_r}|}{M+1} \leq( V^i_{C_r}+\varepsilon_{V} V_{\nu})\\
\end{split}
\label{Eq:EdgeCirclerebelcompareER}
\end{multline}
Lastly, the same evaluation for locating rebel circles are taken place as $C_n$. In general, other parameters (i.e. $L_{C_r}$, $R_{C_r}$ and $Tr$) updates look like the normal circle process except in the estimation, $\beta_{C_r}$ is updating every time with inclusion of  $\frac{\overset{M}{\sum} [\beta_{E_r}+DL_{E_r}]}{M+1}$. 

\begin{figure*}
	\centering
	\includegraphics[width=3.4 in, height = 1.4 in]{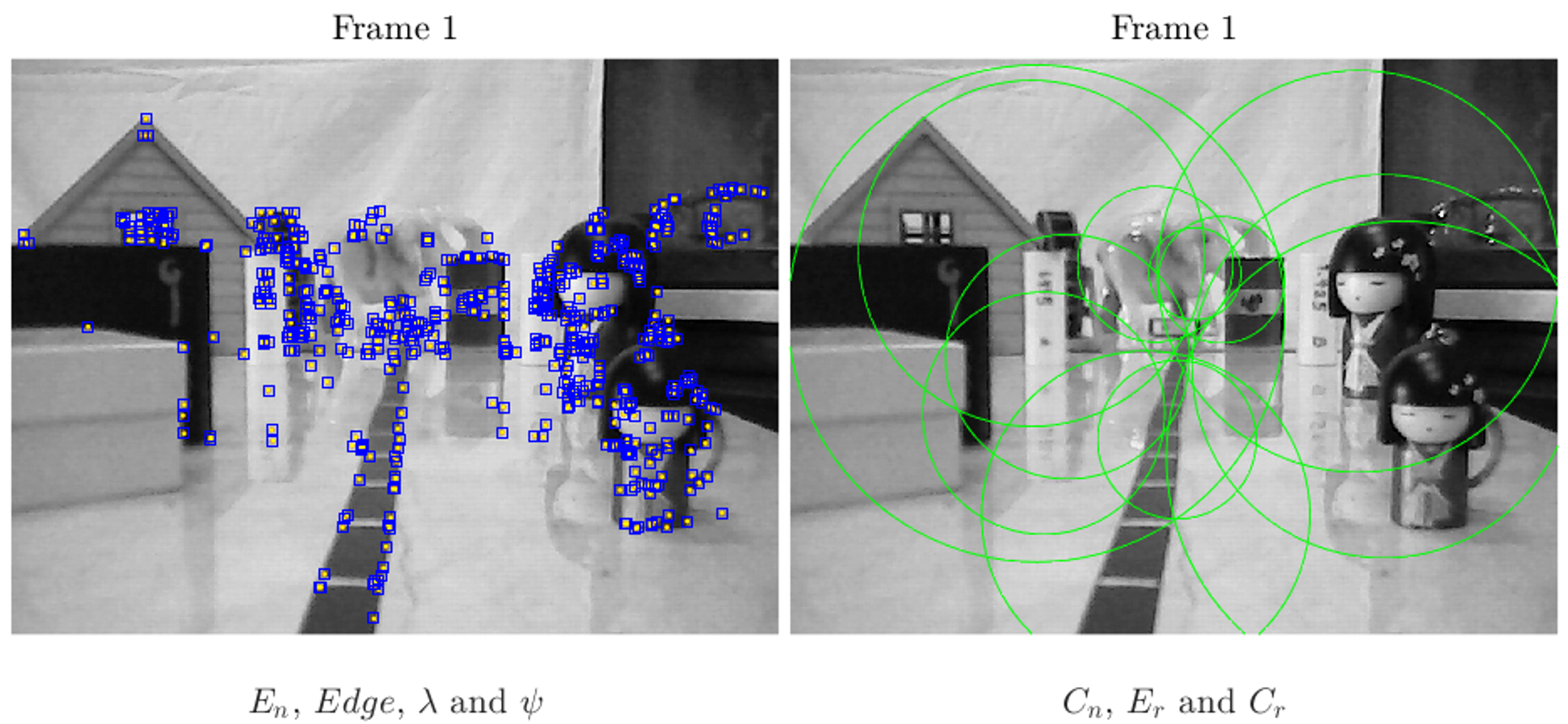}
	\includegraphics[width=3.4 in, height = 1.4 in]{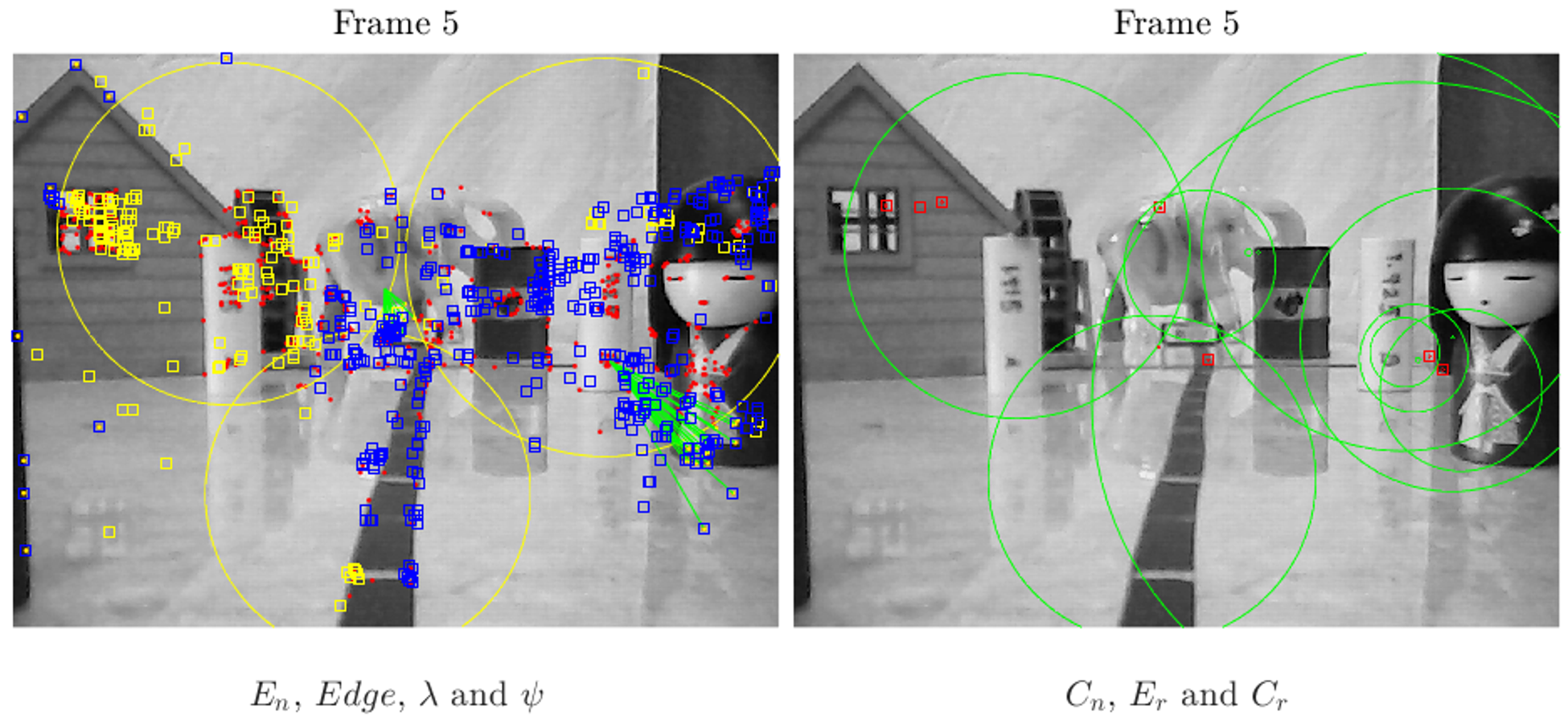}
	\includegraphics[width=3.4 in, height = 1.4 in]{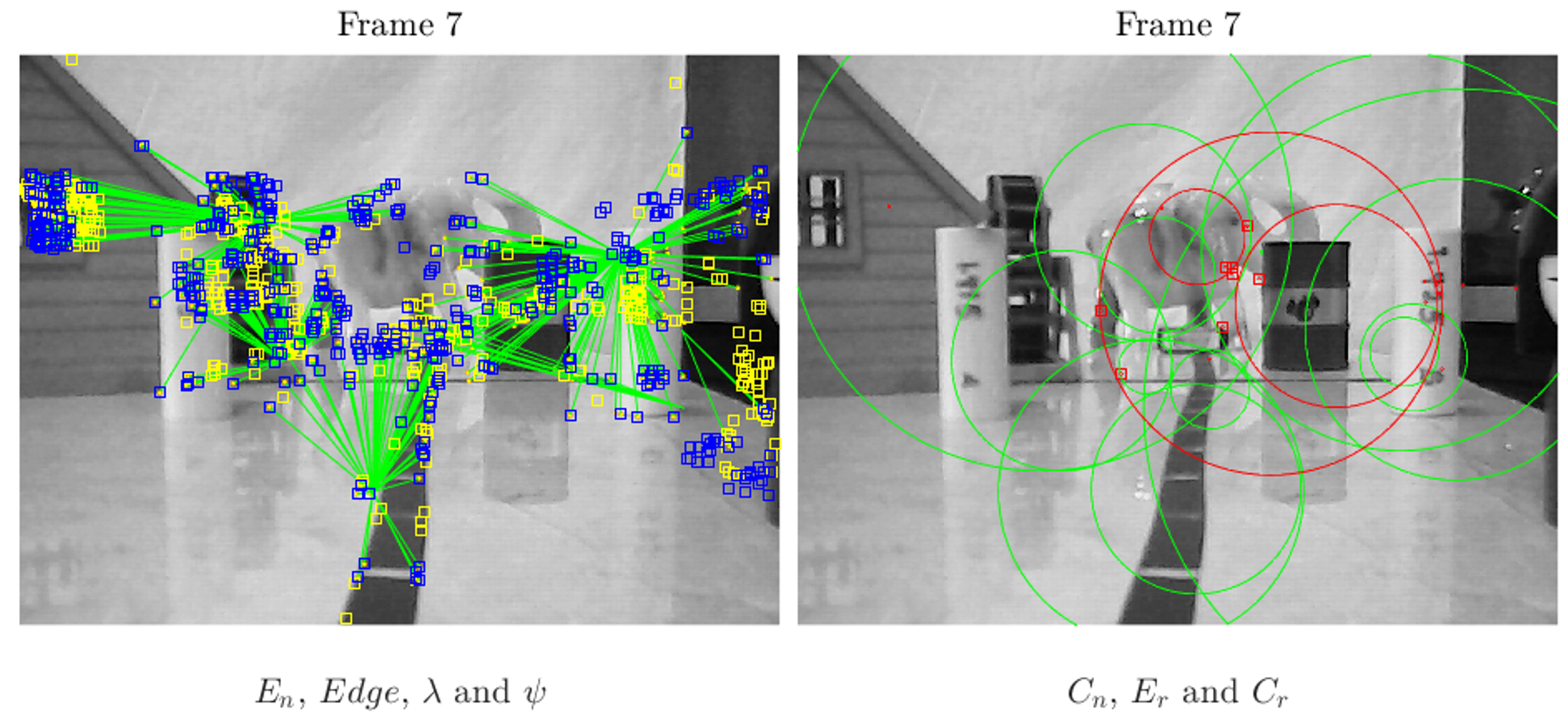}
	\includegraphics[width=3.4 in, height = 1.4 in]{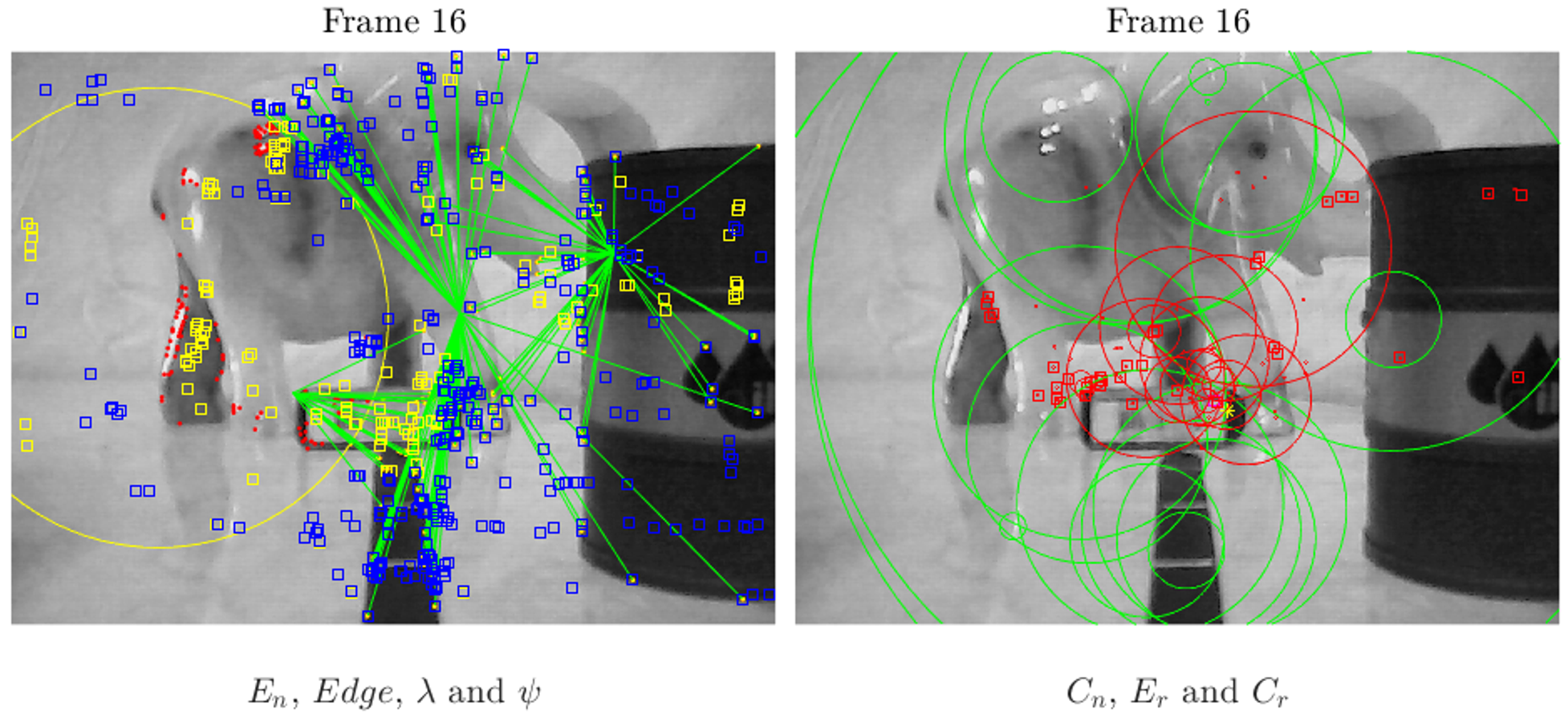}
	\caption{Captured frames of experiment in real-time with camera.}
	\label{Fig:data}
\end{figure*}

\section{Results and Discussion}
In our experiment, we use IMU Samsung sensor integrated with XP HD camera to apply our filter with images resolution of $640 \times 480$ pixels. The study takes place with the worst case scenario where the frame rate is at 1 frame/sec. Although, the geometric filter shows its best performance in at least 3 frame/sec, due to low velocity in vehicle ($V_v$ = $5$ $Cm/s$) and minor acceleration 0.1 $Cm/s^2$, robot can perfectly perform in detecting required edges and passing/incoming objects. In the edge detection technique FAST9 with 25 point threshold is utilized \cite{Rosten2006,4674368}. The environment is chosen with overwhelming landmark properties over 1000 edges spaces. To perfectly apply the algorithm on our trust parameters are determined as $Tr_s$ = $3$, $Tr_{cr}$ = $2$ and $Tr_{max}$ = $5$. Additionally, the angular error related to $\delta^{\nu}_{y-z}$ is 4 pixel on average. The center of the image is in the coordinate location of $320 \times 240$. Boundary layer ($BL$) for initialization of normal edges and detection constant property for rebel edges is 25 pixels. The circling approximated $\varepsilon$ coefficients are $\varepsilon_{\beta}$ = $20^o$ and $\varepsilon_{\nu}$ = $10$ for equation (\ref{EdgeEncompare}) and also  $\varepsilon_{\beta}/5$ and 10 $\varepsilon_{\nu}$ for (\ref{Eq:EdgeCirclenormalcomparee}). For rebel circle coefficients about equation (\ref{Eq:EdgeErcompareER}),  $\varepsilon_{\beta}$ and $\varepsilon_{\nu}$ are $50^o$ and $40$. Additionally, $\varepsilon_{\beta}$ = $10^o$ and  $\varepsilon_{\nu}$ = $1000$ are for last equation (\ref{Eq:EdgeCirclerebelcompareER}). Percentage of involvement for both circling algorithms are 50\%.  The rest of governing parameters in algorithms, are using empty matrices as initial conditions.

The results are demonstrated with four sampled images as Fig. \ref{Fig:data}. The total demonstration has been made available online.$^1$ The presentation has two main alike images. The left side image shows the detected edges with yellow color by FAST9 corner detection algorithm. However, red color points are the omitted ones that detected but is ignored by our filter. Also, the filter latest estimations with our algorithm are shown with blue squares. To present the pure estimation by dynamic equations of motion related to objects, the yellow square graphs are involved. Lastly, to present $\lambda$ and $\psi$, specifically centered connected green lines and yellow circles are used. On the right hand side image, the $C_n$ and $C_r$ are using green and red circles respectively. Lastly, The square graphs are presenting our filters latest $E_r$ estimations in the motion of camera. 

As a highlighted captured frames, in first frame, the filter tries to initialize the scene by recording all detected edges to $E_n$ and preparing normal circles relative to velocity and angle constraints of $C_n$. To compare and sense the change in our experts Frame 5 and 7 are chosen. After, obtaining required trust about certain landmarks with true detection, the $\psi$ is activated to decrease the unwanted calculations. Also, in frame 5, the detected rebel edges results a approximated incoming objects orientations to inform the user. In next frame, it becomes obvious that beside reactivating ignored areas with having automatic edge approximations for those regions, Circle expert tries to locate the previous frames circles which are relative to the objects locations. It is clear that filter let us with minimum frame/rate to detect majority of incoming whole objects or their particular regions. Finally, to one of the last frames (frame 16) with clear incoming objects toward camera, the rebel edge detection with its pure dynamic estimation (purple square), detected current rebel edge (star point) and latest filter estimation (red square) are illustrated. The success of filter can be clearly seen that majority of rebel edges were estimated by filter truly without having clear incoming current frame data. 
\begin{figure}[t!]
	\centering
	\includegraphics[width=2.8 in]{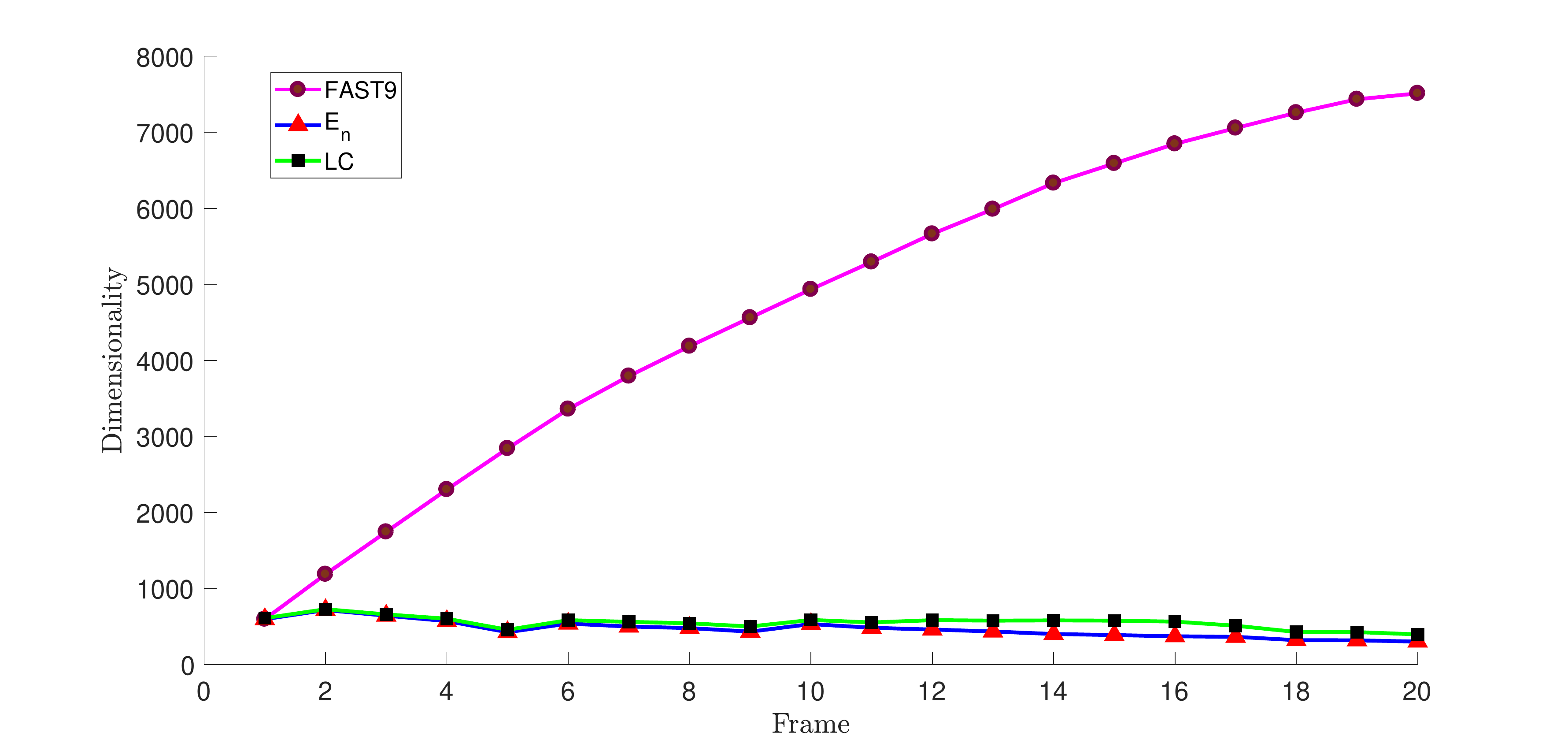}
	\includegraphics[width=2.8 in]{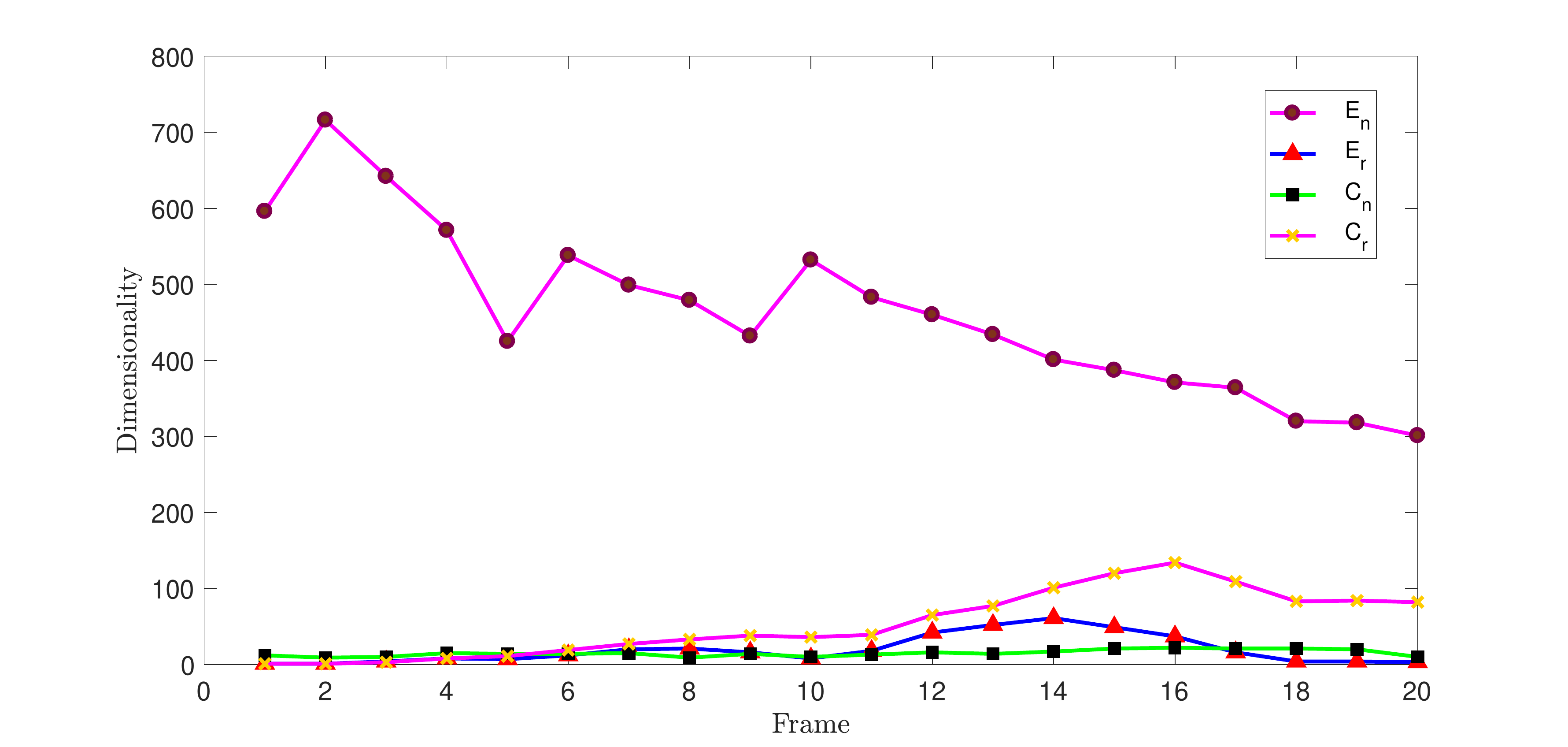}
	\caption{Dimensionality of the LC filter and FAST9 SLAM. Note: the last figure shows LC's each parameter occupying space. }
	\label{Fig:sIZEDImention}
\end{figure}
Lastly, the Fig. \ref{Fig:sIZEDImention} shows that how far the LC filter have computation space in contrast to the FAST9 SLAM. As it is clear, the LC filter dimensionality follows decreasing rate with latest sampled frame at 300. As an interesting fact, although complexity of images with inclusion edges increases, normal circles tries to eliminate overwhelming created edges from normal passing landmarks.

\footnotetext[1]{\href{https://www.dropbox.com/s/m242xdh0uhdb4x5/ICROM_Result.gif?dl=0}{Online available experimental result.}}
\section{Conclusion}
In this paper, we present the integrated geometric LC filter that consists of two experts. The first expert is dealing with collecting the data relative to feedback of circle expert. Detection, learning and tracking all take place inside Circle expert. As an interesting result, beside considerable decrease in matrix sizes in the computations, the model operates smooth follow that let the model act spontaneously with having efficient forgetting and ignoring abilities during continues operations. The ignorance also permit us to avoid any problem during entrance to immense landmark area. 

We are planning to apply error estimation matrix to both experts for operating the filter faster and more accurate in our incoming works. We hope it will be a high potential candidate instead of Kalman Filter. This will help us automate the trust factor perfectly within dynamic scenes. Also, it will be applied on agile non-holonomic robots with rotational plain motions \cite{RollRollerRobotic2016,RollRollerSII2017} to perfectly evaluate the scene properties. Lastly, the filter will be improved as Line-Circle-Square model to completely detect each corresponding objects in the scene. 

\bibliographystyle{IEEEtran}

\bibliography{referencesICROM}

\begin{thebibliography}{10}
\providecommand{\url}[1]{#1}
\csname url@samestyle\endcsname
\providecommand{\newblock}{\relax}
\providecommand{\bibinfo}[2]{#2}
\providecommand{\BIBentrySTDinterwordspacing}{\spaceskip=0pt\relax}
\providecommand{\BIBentryALTinterwordstretchfactor}{4}
\providecommand{\BIBentryALTinterwordspacing}{\spaceskip=\fontdimen2\font plus
\BIBentryALTinterwordstretchfactor\fontdimen3\font minus
  \fontdimen4\font\relax}
\providecommand{\BIBforeignlanguage}[2]{{%
\expandafter\ifx\csname l@#1\endcsname\relax
\typeout{** WARNING: IEEEtran.bst: No hyphenation pattern has been}%
\typeout{** loaded for the language `#1'. Using the pattern for}%
\typeout{** the default language instead.}%
\else
\language=\csname l@#1\endcsname
\fi
#2}}
\providecommand{\BIBdecl}{\relax}
\BIBdecl

\bibitem{Visiononl2014}
H.~Lategahn and C.~Stiller, ``Vision-only localization,'' \emph{IEEE
  Transactions on Intelligent Transportation Systems}, vol.~15, no.~3, pp.
  1246--1257, June 2014.

\bibitem{Xiao2008Facade}
J.~Xiao, T.~Fang, P.~Tan, P.~Zhao, E.~Ofek, and L.~Quan, ``Image-based
  fa\c{C}ade modeling,'' \emph{ACM Trans. Graph.}, vol.~27, no.~5, pp.
  161:1--161:10, Dec. 2008.

\bibitem{Rosten2006}
E.~Rosten and T.~Drummond, ``Machine learning for high-speed corner
  detection,'' in \emph{European Conference on Computer Vision}, Berlin,
  Heidelberg, 2006, pp. 430--443.

\bibitem{Murray2000}
D.~Murray and J.~J. Little, ``Using real-time stereo vision for mobile robot
  navigation,'' \emph{Autonomous Robots}, vol.~8, no.~2, pp. 161--171, 2000.

\bibitem{Sand2008}
P.~Sand and S.~Teller, ``Particle video: Long-range motion estimation using
  point trajectories,'' \emph{International Journal of Computer Vision},
  vol.~80, no.~1, p.~72, 2008.

\bibitem{bibbyeccv08}
C.~Bibby and I.~Reid, ``Robust real-time visual tracking using pixel-wise
  posteriors,'' in \emph{Proceedings of European Conference on Computer
  Vision}, 2008.

\bibitem{PillaiRSS2015}
S.~Pillai and J.~Leonard, ``Monocular slam supported object recognition,'' in
  \emph{Proceedings of Robotics: Science and Systems}, Rome, Italy, July 2015.

\bibitem{VehicleREview2006}
Z.~Sun, G.~Bebis, and R.~Miller, ``On-road vehicle detection: a review,''
  \emph{IEEE Transactions on Pattern Analysis and Machine Intelligence},
  vol.~28, no.~5, pp. 694--711, May 2006.

\bibitem{Tomy2015ICRA}
V.~Lui and T.~Drummond, ``Image based optimisation without global consistency
  for constant time monocular visual slam,'' in \emph{2015 IEEE International
  Conference on Robotics and Automation (ICRA)}, May 2015, pp. 5799--5806.

\bibitem{TommyIROS2015}
D.~Gamage and T.~Drummond, ``Reduced dimensionality extended kalman filter for
  slam in a relative formulation,'' in \emph{2015 IEEE/RSJ International
  Conference on Intelligent Robots and Systems (IROS)}, Sept 2015, pp.
  1365--1372.

\bibitem{EVENTBASECAMERA}
P.~Lichtsteiner, C.~Posch, and T.~Delbruck, ``A 128x128 120 d{B} 15 {$\mu$}s
  latency asynchronous temporal contrast vision sensor,'' \emph{IEEE Journal of
  Solid-State Circuits}, vol.~43, no.~2, pp. 566--576, Feb 2008.

\bibitem{4674368}
E.~Rosten, R.~Porter, and T.~Drummond, ``Faster and better: A machine learning
  approach to corner detection,'' \emph{IEEE Transactions on Pattern Analysis
  and Machine Intelligence}, vol.~32, no.~1, pp. 105--119, Jan 2010.

\bibitem{Bouguet00pyramidalimplementation}
J.~Y. Bouguet, ``Pyramidal implementation of the lucas kanade feature
  tracker,'' \emph{Technical report, Microprocessor Research Labs}, 2000.

\bibitem{TLD2012}
Z.~Kalal, K.~Mikolajczyk, and J.~Matas, ``Tracking-learning-detection,''
  \emph{IEEE Transactions on Pattern Analysis and Machine Intelligence},
  vol.~34, no.~7, pp. 1409--1422, July 2012.

\bibitem{RollRollerRobotic2016}
S.~A. Tafrishi, S.~M. Veres, E.~Esmaeilzadeh, and M.~Svinin, ``Dynamical
  behavior investigation and analysis of novel mechanism for simulated
  spherical robot named {"RollRoller"},'' \emph{ArXiv}, 2016, {A}vilable at
  http://arxiv.org/abs/1610.06218.

\bibitem{RollRollerSII2017}
S.~A. Tafrishi, M.~Svinin, and E.~Esmaeilzadeh, ``Effects of the slope on the
  motion of spherical rollroller robot,'' in \emph{2016 IEEE/SICE International
  Symposium on System Integration (SII)}, Dec 2016, pp. 875--880.

\end{thebibliography}

\end{document}